\documentclass{article}

\usepackage[utf8]{inputenc} 
\usepackage[T1]{fontenc}    
\usepackage{hyperref}       
\usepackage{url}            
\usepackage{booktabs}       
\usepackage{amsfonts}       
\usepackage{nicefrac}       
\usepackage{microtype}      
\usepackage{xcolor}         
\usepackage{lipsum}
\usepackage{appendix}
\usepackage{enumitem}
\usepackage{paralist}
\usepackage{graphicx}
\usepackage{makecell}

\usepackage[accepted]{icml2021}

\newcommand{\mypar}[1]{\textbf{#1}\quad}
\newcommand{\tco}{tCO$_2$e~}

\icmltitlerunning{Toward Foundation Models for Earth Monitoring}

\begin{document}

\twocolumn[
\icmltitle{Toward Foundation Models for Earth Monitoring: \\ Proposal for a Climate Change Benchmark}



\icmlsetsymbol{equal}{*}

\begin{icmlauthorlist}
\icmlauthor{Alexandre Lacoste}{eai}
\icmlauthor{Evan David Sherwin}{stanford}
\icmlauthor{Hannah Kerner}{maryland}
\icmlauthor{Hamed Alemohammad}{radiant}
\icmlauthor{Bj\"orn L\"utjens}{mit}
\icmlauthor{Jeremy Irvin}{stanford}
\icmlauthor{David Dao}{eth}
\icmlauthor{Alex Chang}{eai}
\icmlauthor{Mehmet Gunturkun}{eai}
\icmlauthor{Alexandre Drouin}{eai}
\icmlauthor{Pau Rodriguez}{eai}
\icmlauthor{David Vazquez}{eai}
\end{icmlauthorlist}

\icmlaffiliation{eai}{Element AI / Service Now}
\icmlaffiliation{stanford}{Stanford University}
\icmlaffiliation{radiant}{Radiant Earth Foundation}
\icmlaffiliation{maryland}{University of Maryland}
\icmlaffiliation{mit}{MIT}
\icmlaffiliation{eth}{ETH Zurich}

\icmlcorrespondingauthor{Alexandre Lacoste}{alexandre.lacoste@servicenow.com}

\icmlkeywords{Machine Learning, Climate Change}

\vskip 0.3in
]

\printAffiliationsAndNotice{} 

\begin{abstract}
Recent progress in self-supervision shows that pre-training large neural networks on vast amounts of unsupervised data can lead to impressive increases in generalisation for downstream tasks. Such models, recently coined as \emph{foundation models}, have been transformational to the field of natural language processing. While similar models have also been trained on large corpuses of images, they are not well suited for remote sensing data. To stimulate the development of foundation models for Earth monitoring, we propose to develop a new benchmark comprised of a variety of downstream tasks related to climate change. We believe that this can lead to substantial improvements in many existing applications and facilitate the development of new applications. This proposal is also a call for collaboration with the aim of developing a better evaluation process to mitigate potential downsides of foundation models for Earth monitoring. 
\end{abstract}

\section{Introduction}
Earth monitoring with machine learning-based methods plays an increasing role in climate change mitigation and adaptation as well as climate science~\cite{rolnick2019tackling}. Applications include methane source detection \citep{sheng2020ognet,dileepautomated}, forest carbon quantification~\cite{lutjens2019forest}, deforestation monitoring \cite{Finer2018CombatingDF,deforestation}, flood detection~\cite{meteo-garcia2021floods}, extreme weather prediction~\cite{mcgovern2017weather}, wildfire detection~\cite{piyush2020wildfire}, and crop monitoring~\cite{kerner2020rapid,dado2020high}. Across many of these applications, pre-trained models (e.g., a ResNet trained on ImageNet) are used to increase generalisation performance. Improvement of the pre-trained models is shown to reduce the need for large labelled datasets in some contexts \cite{chen2020simple} and can improve model generalisation outside of the training distribution \cite{hendrycks2019using}. Recent studies exploring the scaling of such pre-trained models found that increasing the size of an unsupervised (or weakly supervised) dataset as well as properly scaling the model led to an even greater increase in performances under various metrics \cite{kaplan2020scaling, radford2021learning}. 

While the training of such large-scale models is usually reserved for industrial research labs with very large computer clusters, the publication of the pre-trained models opens opportunities to the rest of the community. These pre-trained models were recently coined as \emph{foundation models} \cite{bommasani2021opportunities} as they might serve as foundations for sub-fields of machine learning. Specifically, the publication of large pre-trained models like 
BERT \cite{devlin2018bert}, and GPT-3 \cite{brown2020language} led to a paradigm shift in the field of natural language processing (NLP). This inspired a similar shift in the field of computer vision with the release of models like CLIP \cite{radford2021learning} 
and DINO \cite{caron2021emerging}. While CLIP performs well on various types of vision tasks, it is still under-performing on Earth monitoring tasks \cite{radford2021learning}. This is not surprising as it is trained mainly on RGB images taken from a ground perspective, rather than multispectral bands taken from an overhead perspective prevalent in remote sensing data.
This suggests that there is still untapped potential for foundation models to benefit the field Earth monitoring as it has done for NLP and computer vision.

Foundation models also come with downsides. Specifically, large language models are known to amplify and perpetuate biases \cite{bender2021dangers} and have high CO$_2$e emissions associated with their training \cite{strubell2019energy, patterson2021carbon}. Recently, an interdisciplinary group of researchers published a collective work discussing the risks and opportunities of foundation models \cite{bommasani2021opportunities}. This study highlighted that the relevant stakeholders are often not well represented during the design of foundation models. In addition, the increased accessibility of foundation models can lead to the development of unexpected applications with potential positive and negative impacts.
To mitigate potential negative impacts, we suggest an open evaluation procedure early in the process. To this end, we propose a benchmark dataset and evaluation process to facilitate the development of foundation models in Earth monitoring. We will aggregate a collection of downstream tasks such as classification or semantic segmentation to identify ground-based features, provide corresponding labelled datasets, and define a transparent evaluation procedure with open-source code. To highlight the importance of working on climate change, benchmark datasets and tasks will focus on multiple areas related to understanding, mitigating, and adapting to climate change. The advantages of such a benchmark are numerous, as they:
\begin{compactitem} 
    \item stimulate and facilitate the development of foundation models for Earth monitoring,
    \item provide a systematic way of measuring the quality of models for better scientific progress,
    \item provide insights into which pre-trained models work best for specific climate-related tasks, and
    \item preemptively reduce negative impacts of foundation models through an appropriate evaluation procedure.
\end{compactitem}

This work is a proposal and a call to action. We ask the community to engage by proposing suitable datasets, flagging potential concerns, and proposing modifications to the evaluation procedure. In Appendix~\ref{sec:impact}, we review the potential positive and negative societal impacts of this work.

\section{Remote sensing data for self-supervision}
The development of foundation models does not typically rely on a specific dataset for the pre-training phase. The choice of data is part of the design of the model, e.g., a very large corpus of text from the internet \cite{mitchell_never-ending_2018} or pairs of text associated with images from the web \cite{radford2021learning}. To follow this trend, the data for training foundation models will not be provided with the benchmark.
Potential sources of data are listed below.

\mypar{Multispectral with revisits} Data sources such as Sentinel 2 \cite{drusch2012sentinel, esa_sentinel-2_2021} and Landsat 8 \cite{usgs_landsat_2021} provide images in multiple spectral bands with periodic revisits. This yields a 4-dimensional array of structured data (longitude, latitude, wavelength, time) which can be used to perform various forms of self-supervision, e.g., predicting adjacent tiles \cite{jean2019tile2vec} or contrasting the different seasons for the same region \cite{manas2021seasonal}. 

\mypar{Other sensors} Synthetic Aperture Radar (SAR) and terrain elevation are also frequently available and can be matched to other sources of data through geolocalisation \cite{pepin_high-pass_2020}. Such data are complementary to spectral bands and may encourage the model to learn higher-level semantic representations. 

\mypar{Semantic data} Through georeferencing, text-based data such as Wikipedia articles can be linked to satellite images \cite{uzkent2019learning}. It is also possible to join content from non-image data layers like OpenStreetMap \cite{li2020rsi}. By predicting or contrasting information from these sources, the model may learn useful and transferable semantic representations.

\begin{figure*}
    \centering
    \includegraphics[width=0.95\textwidth]{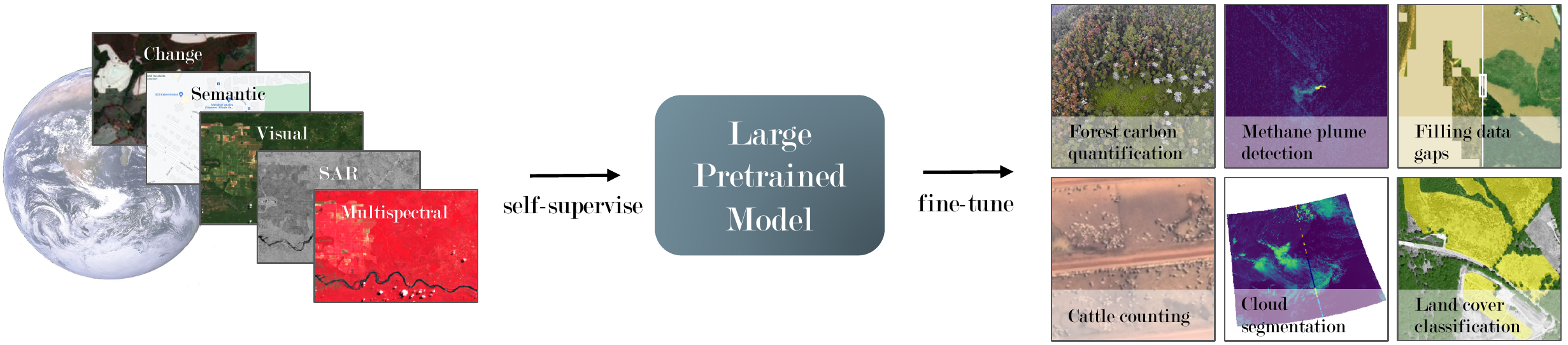}
    \caption{Foundation models encapsulate multimodal data streams through self-supervised training. The trained models can then be fine-tuned for a variety of climate-related remote sensing tasks. Image sources: quantification~\cite{lutjens2019forest}, detection~\cite{jongaramrungruang2021methane}, generation~\cite{lutjens2021virtualsensor}, counting~\cite{laradji2020counting}, segmentation~\cite{zantedeschi2019cumulo}, and multi-class  classification~\cite{pallai2017landcover}.}
    \label{fig:my_label}
\end{figure*}


\section{The Benchmark}
\subsection{Climate Change Downstream Tasks} 
\label{sec:tasks}

The aim is to provide a variety of downstream tasks to evaluate different aspects of foundation models pre-trained on other datasets. To go beyond simple image classification, since it is often not representative of real-world tasks, we include segmentation, regression, and counting tasks. However, for the dataset to be useful in this benchmark, several other criteria need to be met:

\mypar{Not too big} Remote sensing datasets can be comprised of millions of samples totalling terabytes of data. Benchmark datasets should be small enough to easily download onto a personal computer, roughly 100 to a few thousand labelled samples per task. If the license permits it, the dataset can be sub-sampled.

\mypar{Permissive license} Most datasets need to be adapted to fit a conventional machine learning pipeline. In such cases, a permissive license (e.g., Creative Commons) is required. 

\mypar{Multispectral and SAR} One of the main reasons to have a foundation model tailored to remote sensing is to learn how to better interpret multispectral and SAR data. To evaluate their ability to do so, a substantial fraction of the benchmark datasets must contain multispectral and SAR data on tasks that can leverage such information. 

\mypar{Meta information for distribution shift evaluation} We also aim to evaluate model performance under distribution shift, when the model is applied to data from a different distribution than the training data \cite{koh2021wilds}. Of specific interest are downstream tasks in which the training set and the testing set are in different countries. Other variables such as date, sun elevation, and spatial resolution can also provide insightful distribution shift evaluations.

We present current candidate datasets that we are considering for this benchmark in Appendix Table~\ref{tab:datasets}. We encourage the community to contact us to propose additional datasets.


\subsection{Automatic fine-tuning of the model}

To evaluate a pre-trained model, it is common to simply ``probe'' the model, i.e., use the learned representations from the model as the input features to another model \cite{jean2019tile2vec}. 
However, fine-tuning the model to the given task has proven to generalize better and is closer to the needs of practitioners \cite{manas2021seasonal, chen2020simple}. 
Adapting a pre-trained architecture to a variety of types of tasks for fine-tuning comes with significant technical challenges. To this end, we will provide a GitHub codebase with the necessary tools to facilitate and standardise the evaluation procedure. The codebase will provide the following features:

\mypar{Fine-tuning code} To facilitate and standardise fine-tuning, the benchmark will provide code for adapting popular architectures such as ResNet \cite{he2016deep} and Visual Transformer \cite{50650} to the supported types of task such as classification, segmentation and detection.

\mypar{Fine-tuning API} When the pre-trained network is not compatible with existing fine-tuning methods, we encourage the users to submit a pull request to grow the library.

\mypar{Evaluation of representations} Often called probing, this approach does not require fine-tuning. The pre-trained model encode every images of all tasks and predictions are made from the fixed features. This approach requires less computations and is less likely to have compatibility issues.


\subsection{Evaluation Metrics}
We propose to include a variety of metrics to enable rigorous evaluation of the pre-trained models:

\mypar{Task-specific metrics} We propose to report a few metrics that are natural to each task being evaluated, e.g., F1 for classification tasks and mIoU for semantic segmentation.

\mypar{Aggregated metric} For a valid comparison of a pre-trained model across multiple tasks, we will use the pairwise sign test \cite{lacoste2012bayesian}. This simply counts the number of times one model outperforms a baseline and assesses if the difference is significant. When a few strong baselines are compared, Friedman's test \cite{friedman1937use} can be used to provide a more powerful test.

\mypar{Distribution shift} As specified in Section~\ref{sec:tasks}, we are collecting metadata for distribution shift evaluation. This is done by partitioning the train, validation, and test sets of each dataset based on specific values of a selected metadata variable such as country or date. Each partition yields a different evaluation with potentially different insights.

\mypar{Energy efficiency and CO2 equivalent emissions} We will also report energy consumption, and \tco emissions during the benchmarking phase for each model\cite{lacoste2019quantifying,codecarbon}. These emissions are expected to be significantly smaller than that of the pre-training phase, which we do not have access. However, this evaluation will provide a good comparison, highlighting which model is more energy efficient.

\section{Conclusion}

We propose to develop a new benchmark for evaluating foundation models on climate change downstream tasks. This involves adapting a variety of remote sensing datasets to a more conventional machine learning pipeline and providing code for fine-tuning and evaluating on individual tasks. We expect that this benchmark will stimulate the development of new foundation models that could lead to better generalisation on a variety of climate-related downstream tasks and could open up opportunities for new applications.

This proposal is also a call for collaboration. We hope to receive recommendations to include additional public datasets as well as datasets that have not yet been released. We also welcome any recommendations about the evaluation procedure that could improve the validation of foundation models for Earth monitoring and mitigate their potential downsides.

\bibliographystyle{abbrvnat}
\bibliography{references}

\begin{thebibliography}{58}
\providecommand{\natexlab}[1]{#1}
\providecommand{\url}[1]{\texttt{#1}}
\expandafter\ifx\csname urlstyle\endcsname\relax
  \providecommand{\doi}[1]{doi: #1}\else
  \providecommand{\doi}{doi: \begingroup \urlstyle{rm}\Url}\fi

\bibitem[Alemohammad(2021)]{Alemohammad2021}
H.~Alemohammad.
\newblock The case for open-access {ML}-ready geospatial training data.
\newblock In \emph{International Geoscience and Remote Sensing Symposium}.
  IEEE, 2021.

\bibitem[Alemohammad and Booth(2020)]{alemohammad_landcovernet_2020}
H.~Alemohammad and K.~Booth.
\newblock {LandCoverNet}: {A} global benchmark land cover classification
  training dataset.
\newblock \emph{arXiv:2012.03111 [cs]}, Dec. 2020.
\newblock URL \url{http://arxiv.org/abs/2012.03111}.
\newblock arXiv: 2012.03111.

\bibitem[Bender et~al.(2021)Bender, Gebru, McMillan-Major, and
  Shmitchell]{bender2021dangers}
E.~M. Bender, T.~Gebru, A.~McMillan-Major, and S.~Shmitchell.
\newblock On the dangers of stochastic parrots: Can language models be too big?
\newblock In \emph{Proceedings of the 2021 ACM Conference on Fairness,
  Accountability, and Transparency}, pages 610--623, 2021.

\bibitem[Bommasani et~al.(2021)Bommasani, Hudson, Adeli, Altman, Arora, von
  Arx, Bernstein, Bohg, Bosselut, Brunskill,
  et~al.]{bommasani2021opportunities}
R.~Bommasani, D.~A. Hudson, E.~Adeli, R.~Altman, S.~Arora, S.~von Arx, M.~S.
  Bernstein, J.~Bohg, A.~Bosselut, E.~Brunskill, et~al.
\newblock On the opportunities and risks of foundation models.
\newblock \emph{arXiv preprint arXiv:2108.07258}, 2021.

\bibitem[Brown et~al.(2020)Brown, Mann, Ryder, Subbiah, Kaplan, Dhariwal,
  Neelakantan, Shyam, Sastry, Askell, et~al.]{brown2020language}
T.~B. Brown, B.~Mann, N.~Ryder, M.~Subbiah, J.~Kaplan, P.~Dhariwal,
  A.~Neelakantan, P.~Shyam, G.~Sastry, A.~Askell, et~al.
\newblock Language models are few-shot learners.
\newblock \emph{arXiv preprint arXiv:2005.14165}, 2020.

\bibitem[Burke et~al.(2021)Burke, Driscoll, Lobell, and Ermon]{burke2021using}
M.~Burke, A.~Driscoll, D.~B. Lobell, and S.~Ermon.
\newblock Using satellite imagery to understand and promote sustainable
  development.
\newblock \emph{Science}, 371\penalty0 (6535), 2021.

\bibitem[Caron et~al.(2021)Caron, Touvron, Misra, J{\'e}gou, Mairal,
  Bojanowski, and Joulin]{caron2021emerging}
M.~Caron, H.~Touvron, I.~Misra, H.~J{\'e}gou, J.~Mairal, P.~Bojanowski, and
  A.~Joulin.
\newblock Emerging properties in self-supervised vision transformers.
\newblock \emph{arXiv preprint arXiv:2104.14294}, 2021.

\bibitem[Chen et~al.(2020)Chen, Kornblith, Norouzi, and Hinton]{chen2020simple}
T.~Chen, S.~Kornblith, M.~Norouzi, and G.~Hinton.
\newblock A simple framework for contrastive learning of visual
  representations.
\newblock In \emph{International conference on machine learning}, pages
  1597--1607. PMLR, 2020.

\bibitem[Chiu et~al.(2020)Chiu, Xu, Wei, Huang, Schwing, Brunner, Khachatrian,
  Karapetyan, Dozier, Rose, Wilson, Tudor, Hovakimyan, Huang, and
  Shi]{chiu_agriculture-vision_2020}
M.~T. Chiu, X.~Xu, Y.~Wei, Z.~Huang, A.~G. Schwing, R.~Brunner, H.~Khachatrian,
  H.~Karapetyan, I.~Dozier, G.~Rose, D.~Wilson, A.~Tudor, N.~Hovakimyan, T.~S.
  Huang, and H.~Shi.
\newblock Agriculture-{Vision}: {A} {Large} {Aerial} {Image} {Database} for
  {Agricultural} {Pattern} {Analysis}.
\newblock In \emph{2020 {IEEE}/{CVF} {Conference} on {Computer} {Vision} and
  {Pattern} {Recognition} ({CVPR})}, pages 2825--2835, Seattle, WA, USA, June
  2020. IEEE.
\newblock ISBN 978-1-72817-168-5.
\newblock \doi{10.1109/CVPR42600.2020.00290}.
\newblock URL \url{https://ieeexplore.ieee.org/document/9157397/}.

\bibitem[Dado et~al.(2020)Dado, Deines, Patel, Liang, and Lobell]{dado2020high}
W.~T. Dado, J.~M. Deines, R.~Patel, S.-Z. Liang, and D.~B. Lobell.
\newblock High-resolution soybean yield mapping across the us midwest using
  subfield harvester data.
\newblock \emph{Remote Sensing}, 12\penalty0 (21):\penalty0 3471, 2020.

\bibitem[Dao et~al.()Dao, Cang, Fung, Zhang, Pawlowski, Gonzales, Beglinger,
  and Zhang]{deforestation}
D.~Dao, C.~Cang, C.~Fung, M.~Zhang, N.~Pawlowski, R.~Gonzales, N.~Beglinger,
  and C.~Zhang.
\newblock Gainforest: Scaling climate finance for forest conservation using
  interpretable machine learning on satellite imagery.

\bibitem[Devlin et~al.(2018)Devlin, Chang, Lee, and Toutanova]{devlin2018bert}
J.~Devlin, M.-W. Chang, K.~Lee, and K.~Toutanova.
\newblock Bert: Pre-training of deep bidirectional transformers for language
  understanding.
\newblock \emph{arXiv preprint arXiv:1810.04805}, 2018.

\bibitem[Dileep et~al.(2020)Dileep, Zimmerle, Beveridge, and
  Vaughn]{dileepautomated}
S.~Dileep, D.~Zimmerle, J.~R. Beveridge, and T.~Vaughn.
\newblock Automated identification of oil field features using cnns.
\newblock 2020.

\bibitem[Drusch et~al.(2012)Drusch, Del~Bello, Carlier, Colin, Fernandez,
  Gascon, Hoersch, Isola, Laberinti, Martimort, et~al.]{drusch2012sentinel}
M.~Drusch, U.~Del~Bello, S.~Carlier, O.~Colin, V.~Fernandez, F.~Gascon,
  B.~Hoersch, C.~Isola, P.~Laberinti, P.~Martimort, et~al.
\newblock Sentinel-2: Esa's optical high-resolution mission for gmes
  operational services.
\newblock \emph{Remote sensing of Environment}, 120:\penalty0 25--36, 2012.

\bibitem[Duren et~al.(2019)Duren, Thorpe, Foster, Rafiq, Hopkins, Yadav, Bue,
  Thompson, Conley, Colombi, Frankenberg, McCubbin, Eastwood, Falk, Herner,
  Croes, Green, and Miller]{duren_californias_2019}
R.~M. Duren, A.~K. Thorpe, K.~T. Foster, T.~Rafiq, F.~M. Hopkins, V.~Yadav,
  B.~D. Bue, D.~R. Thompson, S.~Conley, N.~K. Colombi, C.~Frankenberg, I.~B.
  McCubbin, M.~L. Eastwood, M.~Falk, J.~D. Herner, B.~E. Croes, R.~O. Green,
  and C.~E. Miller.
\newblock California's methane super-emitters.
\newblock \emph{Nature}, 575\penalty0 (7781):\penalty0 180--184, Nov. 2019.
\newblock ISSN 0028-0836, 1476-4687.
\newblock \doi{10.1038/s41586-019-1720-3}.
\newblock URL \url{http://www.nature.com/articles/s41586-019-1720-3}.

\bibitem[EPA(2017)]{epa_greenhouse_2017}
EPA.
\newblock Greenhouse {Gas} {Emissions}: {Understanding} {Global} {Warming}
  {Potentials}.
\newblock Technical report, US Environmental Protection Agency, Feb. 2017.
\newblock URL
  \url{https://www.epa.gov/ghgemissions/understanding-global-warming-potentials}.

\bibitem[ESA(2021)]{esa_sentinel-2_2021}
ESA.
\newblock Sentinel-2.
\newblock Technical report, European Space Agency, Paris, France, 2021.
\newblock URL \url{https://sentinel.esa.int/web/sentinel/missions/sentinel-2}.

\bibitem[Finer et~al.(2018)Finer, Novoa, Weisse, Petersen, Mascaro, Souto,
  Stearns, and Martinez]{Finer2018CombatingDF}
M.~Finer, S.~Novoa, M.~Weisse, R.~Petersen, J.~Mascaro, T.~Souto, F.~Stearns,
  and R.~Martinez.
\newblock Combating deforestation: From satellite to intervention.
\newblock \emph{Science}, 360:\penalty0 1303 -- 1305, 2018.

\bibitem[Friedman(1937)]{friedman1937use}
M.~Friedman.
\newblock The use of ranks to avoid the assumption of normality implicit in the
  analysis of variance.
\newblock \emph{Journal of the american statistical association}, 32\penalty0
  (200):\penalty0 675--701, 1937.

\bibitem[He et~al.(2016)He, Zhang, Ren, and Sun]{he2016deep}
K.~He, X.~Zhang, S.~Ren, and J.~Sun.
\newblock Deep residual learning for image recognition.
\newblock In \emph{Proceedings of the IEEE conference on computer vision and
  pattern recognition}, pages 770--778, 2016.

\bibitem[Hendrycks et~al.(2019)Hendrycks, Mazeika, Kadavath, and
  Song]{hendrycks2019using}
D.~Hendrycks, M.~Mazeika, S.~Kadavath, and D.~Song.
\newblock Using self-supervised learning can improve model robustness and
  uncertainty.
\newblock \emph{arXiv preprint arXiv:1906.12340}, 2019.

\bibitem[Jain et~al.(2020)Jain, Coogan, Subramanian, Crowley, Taylor, and
  Flannigan]{piyush2020wildfire}
P.~Jain, S.~C. Coogan, S.~G. Subramanian, M.~Crowley, S.~Taylor, and M.~D.
  Flannigan.
\newblock A review of machine learning applications in wildfire science and
  management.
\newblock \emph{Environmental Reviews}, 28\penalty0 (4):\penalty0 478--505,
  2020.

\bibitem[Jean et~al.(2019)Jean, Wang, Samar, Azzari, Lobell, and
  Ermon]{jean2019tile2vec}
N.~Jean, S.~Wang, A.~Samar, G.~Azzari, D.~Lobell, and S.~Ermon.
\newblock Tile2vec: Unsupervised representation learning for spatially
  distributed data.
\newblock In \emph{Proceedings of the AAAI Conference on Artificial
  Intelligence}, volume~33, pages 3967--3974, 2019.

\bibitem[Johnson et~al.(2021)Johnson, Wlazlo, Keys, Desai, Wetherley, Calvert,
  and Berman]{johnson_airborne_2021}
F.~Johnson, A.~Wlazlo, R.~Keys, V.~Desai, E.~Wetherley, R.~Calvert, and
  E.~Berman.
\newblock Airborne methane surveys pay for themselves: {An} economic case study
  of increased revenue from emissions control.
\newblock preprint, Environmental Monitoring, July 2021.
\newblock URL \url{http://eartharxiv.org/repository/view/2532/}.

\bibitem[Jongaramrungruang et~al.(2021)Jongaramrungruang, Frankenberg, Thorpe,
  and Matheou]{jongaramrungruang2021methane}
S.~Jongaramrungruang, C.~Frankenberg, A.~K. Thorpe, and G.~Matheou.
\newblock Methanet - an ai-driven approach to quantifying methane point-source
  emission from high-resolution 2-d plume imagery.
\newblock \emph{ICML Workshop on Tackling Climate Change with AI}, 2021.

\bibitem[Kaplan et~al.(2020)Kaplan, McCandlish, Henighan, Brown, Chess, Child,
  Gray, Radford, Wu, and Amodei]{kaplan2020scaling}
J.~Kaplan, S.~McCandlish, T.~Henighan, T.~B. Brown, B.~Chess, R.~Child,
  S.~Gray, A.~Radford, J.~Wu, and D.~Amodei.
\newblock Scaling laws for neural language models.
\newblock \emph{arXiv preprint arXiv:2001.08361}, 2020.

\bibitem[Kerner et~al.(2020)Kerner, Tseng, Becker-Reshef, Nakalembe, Barker,
  Munshell, Paliyam, and Hosseini]{kerner2020rapid}
H.~Kerner, G.~Tseng, I.~Becker-Reshef, C.~Nakalembe, B.~Barker, B.~Munshell,
  M.~Paliyam, and M.~Hosseini.
\newblock Rapid response crop maps in data sparse regions.
\newblock \emph{arXiv preprint arXiv:2006.16866}, 2020.

\bibitem[Koh et~al.(2021)Koh, Sagawa, Xie, Zhang, Balsubramani, Hu, Yasunaga,
  Phillips, Gao, Lee, et~al.]{koh2021wilds}
P.~W. Koh, S.~Sagawa, S.~M. Xie, M.~Zhang, A.~Balsubramani, W.~Hu, M.~Yasunaga,
  R.~L. Phillips, I.~Gao, T.~Lee, et~al.
\newblock Wilds: A benchmark of in-the-wild distribution shifts.
\newblock In \emph{International Conference on Machine Learning}, pages
  5637--5664. PMLR, 2021.

\bibitem[Kolesnikov et~al.(2021)Kolesnikov, Dosovitskiy, Weissenborn, Heigold,
  Uszkoreit, Beyer, Minderer, Dehghani, Houlsby, Gelly, Unterthiner, and
  Zhai]{50650}
A.~Kolesnikov, A.~Dosovitskiy, D.~Weissenborn, G.~Heigold, J.~Uszkoreit,
  L.~Beyer, M.~Minderer, M.~Dehghani, N.~Houlsby, S.~Gelly, T.~Unterthiner, and
  X.~Zhai.
\newblock An image is worth 16x16 words: Transformers for image recognition at
  scale.
\newblock 2021.

\bibitem[Lacoste et~al.(2012)Lacoste, Laviolette, and
  Marchand]{lacoste2012bayesian}
A.~Lacoste, F.~Laviolette, and M.~Marchand.
\newblock Bayesian comparison of machine learning algorithms on single and
  multiple datasets.
\newblock In \emph{Artificial Intelligence and Statistics}, pages 665--675.
  PMLR, 2012.

\bibitem[Lacoste et~al.(2019)Lacoste, Luccioni, Schmidt, and
  Dandres]{lacoste2019quantifying}
A.~Lacoste, A.~Luccioni, V.~Schmidt, and T.~Dandres.
\newblock Quantifying the carbon emissions of machine learning.
\newblock \emph{arXiv preprint arXiv:1910.09700}, 2019.

\bibitem[Laradji et~al.(2020)Laradji, Rodriguez, Kalaitzis, Vazquez, Young,
  Davey, and Lacoste]{laradji2020counting}
I.~Laradji, P.~Rodriguez, F.~Kalaitzis, D.~Vazquez, R.~Young, E.~Davey, and
  A.~Lacoste.
\newblock Counting cows: Tracking illegal cattle ranching from high-resolution
  satellite imagery.
\newblock \emph{arXiv preprint arXiv:2011.07369}, 2020.

\bibitem[Li et~al.(2020)Li, Dou, Tao, Wu, Chen, Peng, Deng, and
  Zhao]{li2020rsi}
H.~Li, X.~Dou, C.~Tao, Z.~Wu, J.~Chen, J.~Peng, M.~Deng, and L.~Zhao.
\newblock Rsi-cb: A large-scale remote sensing image classification benchmark
  using crowdsourced data.
\newblock \emph{Sensors}, 20\penalty0 (6):\penalty0 1594, 2020.

\bibitem[L\"utjens et~al.(2019)L\"utjens, Liebenwein, and
  Kramer]{lutjens2019forest}
B.~L\"utjens, L.~Liebenwein, and K.~Kramer.
\newblock Machine learning-based estimation of forest carbon stocks to increase
  transparency of forest preservation efforts.
\newblock \emph{2019 NeurIPS Workshop on Tackling Climate Change with AI
  (CCAI)}, 2019.

\bibitem[L\"utjens et~al.(2021)L\"utjens, Leshchinskiy, Requena-Mesa, Chishtie,
  D\'iaz-Rodr\'iguez, Boulais, Sankaranarayanan, Pina, Gal, Raissi, Lavin, and
  Newman]{lutjens2021virtualsensor}
B.~L\"utjens, B.~Leshchinskiy, C.~Requena-Mesa, F.~Chishtie,
  N.~D\'iaz-Rodr\'iguez, O.~Boulais, A.~Sankaranarayanan, A.~Pina, Y.~Gal,
  C.~Raissi, A.~Lavin, and D.~Newman.
\newblock Physically-consistent generative adversarial networks for coastal
  flood visualization.
\newblock \emph{ICML Workshop on AI for Modeling Oceans and Climate Change
  (AIMOCC)}, 2021.

\bibitem[Ma et~al.(2019)Ma, Liu, Zhang, Ye, Yin, and Johnson]{ma2019deep}
L.~Ma, Y.~Liu, X.~Zhang, Y.~Ye, G.~Yin, and B.~A. Johnson.
\newblock Deep learning in remote sensing applications: A meta-analysis and
  review.
\newblock \emph{ISPRS journal of photogrammetry and remote sensing},
  152:\penalty0 166--177, 2019.

\bibitem[Ma{\~n}as et~al.(2021)Ma{\~n}as, Lacoste, Giro-i Nieto, Vazquez, and
  Rodriguez]{manas2021seasonal}
O.~Ma{\~n}as, A.~Lacoste, X.~Giro-i Nieto, D.~Vazquez, and P.~Rodriguez.
\newblock Seasonal contrast: Unsupervised pre-training from uncurated remote
  sensing data.
\newblock \emph{arXiv preprint arXiv:2103.16607}, 2021.

\bibitem[Maskey et~al.(2020{\natexlab{a}})Maskey, Alemohammad, Murphy, and
  Ramachandran]{maskey2020advancing}
M.~Maskey, H.~Alemohammad, K.~Murphy, and R.~Ramachandran.
\newblock Advancing ai for earth science: A data systems perspective.
\newblock \emph{Eos}, 101, 2020{\natexlab{a}}.

\bibitem[Maskey et~al.(2020{\natexlab{b}})Maskey, Ramachandran,
  Ramasubramanian, Gurung, Freitag, Kaulfus, Bollinger, Cecil, and
  Miller]{Maskey2020}
M.~Maskey, R.~Ramachandran, M.~Ramasubramanian, I.~Gurung, B.~Freitag,
  A.~Kaulfus, D.~Bollinger, D.~J. Cecil, and J.~Miller.
\newblock Deepti: Deep-learning-based tropical cyclone intensity estimation
  system.
\newblock \emph{IEEE Journal of Selected Topics in Applied Earth Observations
  and Remote Sensing}, 13:\penalty0 4271--4281, 2020{\natexlab{b}}.
\newblock \doi{10.1109/JSTARS.2020.3011907}.

\bibitem[Mateo-Garcia et~al.(2021)Mateo-Garcia, Veitch-Michaelis, Smith, Oprea,
  Schumann, Gal, Baydin, and Backes]{meteo-garcia2021floods}
G.~Mateo-Garcia, J.~Veitch-Michaelis, L.~Smith, S.~V. Oprea, G.~Schumann,
  Y.~Gal, A.~G. Baydin, and D.~Backes.
\newblock Towards global flood mapping onboard low cost satellites with machine
  learning.
\newblock \emph{Scientific Reports}, 11, 2021.

\bibitem[McGovern et~al.(2017)McGovern, Elmore, Gagne, Haupt, Karstens,
  Lagerquist, Smith, and Williams]{mcgovern2017weather}
A.~McGovern, K.~L. Elmore, D.~J. Gagne, S.~E. Haupt, C.~D. Karstens,
  R.~Lagerquist, T.~Smith, and J.~K. Williams.
\newblock Using artificial intelligence to improve real-time decision-making
  for high-impact weather.
\newblock \emph{Bulletin of the American Meteorological Society}, 98\penalty0
  (10), 2017.

\bibitem[Mitchell et~al.(2018)Mitchell, Cohen, Hruschka, Talukdar, Yang,
  Betteridge, Carlson, Dalvi, Gardner, Kisiel, Krishnamurthy, Lao, Mazaitis,
  Mohamed, Nakashole, Platanios, Ritter, Samadi, Settles, Wang, Wijaya, Gupta,
  Chen, Saparov, Greaves, and Welling]{mitchell_never-ending_2018}
T.~Mitchell, W.~Cohen, E.~Hruschka, P.~Talukdar, B.~Yang, J.~Betteridge,
  A.~Carlson, B.~Dalvi, M.~Gardner, B.~Kisiel, J.~Krishnamurthy, N.~Lao,
  K.~Mazaitis, T.~Mohamed, N.~Nakashole, E.~Platanios, A.~Ritter, M.~Samadi,
  B.~Settles, R.~Wang, D.~Wijaya, A.~Gupta, X.~Chen, A.~Saparov, M.~Greaves,
  and J.~Welling.
\newblock Never-ending learning.
\newblock \emph{Communications of the ACM}, 61\penalty0 (5):\penalty0 103--115,
  Apr. 2018.
\newblock ISSN 0001-0782, 1557-7317.
\newblock \doi{10.1145/3191513}.
\newblock URL \url{https://dl.acm.org/doi/10.1145/3191513}.

\bibitem[Pallai and Wesson(2017)]{pallai2017landcover}
C.~Pallai and K.~Wesson.
\newblock Chesapeake bay program partnership high-resolution land cover
  classification accuracy assessment methodology, 2017.
\newblock URL
  \url{https://chesapeakeconservancy.org/wp-content/uploads/2017/01/Chesapeake_Conservancy_Accuracy_Assessment_Methodology.pdf}.

\bibitem[Patterson et~al.(2021)Patterson, Gonzalez, Le, Liang, Munguia,
  Rothchild, So, Texier, and Dean]{patterson2021carbon}
D.~Patterson, J.~Gonzalez, Q.~Le, C.~Liang, L.-M. Munguia, D.~Rothchild, D.~So,
  M.~Texier, and J.~Dean.
\newblock Carbon emissions and large neural network training.
\newblock \emph{arXiv preprint arXiv:2104.10350}, 2021.

\bibitem[Pepin et~al.(2020)Pepin, Zebker, and Ellsworth]{pepin_high-pass_2020}
K.~Pepin, H.~A. Zebker, and W.~Ellsworth.
\newblock High-{Pass} {Filters} to {Reduce} the {Effects} of {Broad}
  {Atmospheric} {Contributions} in {Sbas} {Inversions}: {A} {Case} {Study} in
  the {Delaware} {Basin}.
\newblock In \emph{{IGARSS} 2020 - 2020 {IEEE} {International} {Geoscience} and
  {Remote} {Sensing} {Symposium}}, pages 1030--1033, Waikoloa, HI, USA, Sept.
  2020. IEEE.
\newblock ISBN 978-1-72816-374-1.
\newblock \doi{10.1109/IGARSS39084.2020.9324656}.
\newblock URL \url{https://ieeexplore.ieee.org/document/9324656/}.

\bibitem[Radford et~al.(2021)Radford, Kim, Hallacy, Ramesh, Goh, Agarwal,
  Sastry, Askell, Mishkin, Clark, et~al.]{radford2021learning}
A.~Radford, J.~W. Kim, C.~Hallacy, A.~Ramesh, G.~Goh, S.~Agarwal, G.~Sastry,
  A.~Askell, P.~Mishkin, J.~Clark, et~al.
\newblock Learning transferable visual models from natural language
  supervision.
\newblock \emph{arXiv preprint arXiv:2103.00020}, 2021.

\bibitem[Rambour et~al.(2020)Rambour, Audebert, Koeniguer, Le~Saux, Crucianu,
  and Datcu]{rambour_flood_2020}
C.~Rambour, N.~Audebert, E.~Koeniguer, B.~Le~Saux, M.~Crucianu, and M.~Datcu.
\newblock Flood detection in time series of optical and {SAR} images.
\newblock \emph{The International Archives of the Photogrammetry, Remote
  Sensing and Spatial Information Sciences}, XLIII-B2-2020:\penalty0
  1343--1346, Aug. 2020.
\newblock ISSN 2194-9034.
\newblock \doi{10.5194/isprs-archives-XLIII-B2-2020-1343-2020}.
\newblock URL
  \url{https://www.int-arch-photogramm-remote-sens-spatial-inf-sci.net/XLIII-B2-2020/1343/2020/}.

\bibitem[Rausch et~al.(2020)Rausch, Mayer, Arlt, Gust, Staudt, Weinhardt,
  Neumann, and Rajagopal]{rausch2020enriched}
B.~Rausch, K.~Mayer, M.-L. Arlt, G.~Gust, P.~Staudt, C.~Weinhardt, D.~Neumann,
  and R.~Rajagopal.
\newblock An enriched automated pv registry: Combining image recognition and 3d
  building data.
\newblock \emph{arXiv preprint arXiv:2012.03690}, 2020.

\bibitem[Rolnick et~al.(2019)Rolnick, Donti, Kaack, Kochanski, Lacoste,
  Sankaran, Ross, Milojevic-Dupont, Jaques, Waldman-Brown,
  et~al.]{rolnick2019tackling}
D.~Rolnick, P.~L. Donti, L.~H. Kaack, K.~Kochanski, A.~Lacoste, K.~Sankaran,
  A.~S. Ross, N.~Milojevic-Dupont, N.~Jaques, A.~Waldman-Brown, et~al.
\newblock Tackling climate change with machine learning.
\newblock \emph{arXiv preprint arXiv:1906.05433}, 2019.

\bibitem[Ross et~al.(2019)Ross, Topp, Appling, Yang, Kuhn, Butman, Simard, and
  Pavelsky]{ross_aquasat_2019}
M.~R.~V. Ross, S.~N. Topp, A.~P. Appling, X.~Yang, C.~Kuhn, D.~Butman,
  M.~Simard, and T.~M. Pavelsky.
\newblock {AquaSat}: {A} {Data} {Set} to {Enable} {Remote} {Sensing} of {Water}
  {Quality} for {Inland} {Waters}.
\newblock \emph{Water Resources Research}, 55\penalty0 (11):\penalty0
  10012--10025, Nov. 2019.
\newblock ISSN 0043-1397, 1944-7973.
\newblock \doi{10.1029/2019WR024883}.
\newblock URL \url{https://onlinelibrary.wiley.com/doi/10.1029/2019WR024883}.

\bibitem[Schmidt et~al.(2021)Schmidt, Goyal, Joshi, Feld, Conell, Laskaris,
  Blank, Wilson, Friedler, and Luccioni]{codecarbon}
V.~Schmidt, K.~Goyal, A.~Joshi, B.~Feld, L.~Conell, N.~Laskaris, D.~Blank,
  J.~Wilson, S.~Friedler, and S.~Luccioni.
\newblock {CodeCarbon: Estimate and Track Carbon Emissions from Machine
  Learning Computing}.
\newblock 2021.
\newblock \doi{10.5281/zenodo.4658424}.

\bibitem[Schwartz et~al.(2020)Schwartz, Dodge, Smith, and
  Etzioni]{schwartz2020green}
R.~Schwartz, J.~Dodge, N.~A. Smith, and O.~Etzioni.
\newblock Green ai.
\newblock \emph{Communications of the ACM}, 63\penalty0 (12):\penalty0 54--63,
  2020.

\bibitem[Sheng et~al.(2020)Sheng, Irvin, Munukutla, Zhang, Cross, Story,
  Rustowicz, Elsworth, Yang, Omara, et~al.]{sheng2020ognet}
H.~Sheng, J.~Irvin, S.~Munukutla, S.~Zhang, C.~Cross, K.~Story, R.~Rustowicz,
  C.~Elsworth, Z.~Yang, M.~Omara, et~al.
\newblock Ognet: Towards a global oil and gas infrastructure database using
  deep learning on remotely sensed imagery.
\newblock \emph{arXiv preprint arXiv:2011.07227}, 2020.

\bibitem[Strubell et~al.(2019)Strubell, Ganesh, and
  McCallum]{strubell2019energy}
E.~Strubell, A.~Ganesh, and A.~McCallum.
\newblock Energy and policy considerations for deep learning in nlp.
\newblock In \emph{Proceedings of the 57th Annual Meeting of the Association
  for Computational Linguistics}, pages 3645--3650, 2019.

\bibitem[USGS(2021)]{usgs_landsat_2021}
USGS.
\newblock Landsat 8.
\newblock Technical report, United States Geological Survey, Reston, Virginia,
  USA, 2021.
\newblock URL
  \url{https://www.usgs.gov/core-science-systems/nli/landsat/landsat-8?qt-science_support_page_related_con=0#qt-science_support_page_related_con}.

\bibitem[Uzkent et~al.(2019)Uzkent, Sheehan, Meng, Tang, Burke, Lobell, and
  Ermon]{uzkent2019learning}
B.~Uzkent, E.~Sheehan, C.~Meng, Z.~Tang, M.~Burke, D.~Lobell, and S.~Ermon.
\newblock Learning to interpret satellite images using wikipedia.
\newblock In \emph{Proceedings of the Twenty-Eighth International Joint
  Conference on Artificial Intelligence}, 2019.

\bibitem[Zantedeschi et~al.(2019)Zantedeschi, Falasca, Douglas, Strange,
  Kusner, and Watson-Parris]{zantedeschi2019cumulo}
V.~Zantedeschi, F.~Falasca, A.~Douglas, R.~Strange, M.~J. Kusner, and
  D.~Watson-Parris.
\newblock Cumulo: A dataset for learning cloud classes.
\newblock \emph{arXiv preprint arXiv:1911.04227}, 2019.

\bibitem[Zhu et~al.(2017)Zhu, Tuia, Mou, Xia, Zhang, Xu, and
  Fraundorfer]{zhu2017deep}
X.~X. Zhu, D.~Tuia, L.~Mou, G.-S. Xia, L.~Zhang, F.~Xu, and F.~Fraundorfer.
\newblock Deep learning in remote sensing: A comprehensive review and list of
  resources.
\newblock \emph{IEEE Geoscience and Remote Sensing Magazine}, 5\penalty0
  (4):\penalty0 8--36, 2017.

\end{thebibliography}

\appendix

\section{Societal Impact of Foundation Models for Earth Monitoring} \label{sec:impact}

Remote sensing and Earth monitoring have been transformational in the past decades. Applications include military, insurance, market forecasting, climate science, and more. Much of this impact is not directly attributed to deep learning nor large pre-trained networks and its review extends beyond the scope of this section. In this section, our focus is on the impact of bringing foundation models to Earth monitoring.

\subsection{Climate mitigation and adaptation}
Machine learning on remote sensing data is widely used to develop solutions for a variety of problems relevant to climate change \cite{burke2021using,rolnick2019tackling,zhu2017deep,ma2019deep}. The vast majority of these solutions are built by curating datasets for a specific task and require significant resources to develop. Furthermore, the solutions are often tailored to specific regions as extending  approaches to new geographies remains a significant challenge, primarily due to the lack of labeled data \cite{zhu2017deep}. Less-economically developed regions of the world are no less susceptible to the impacts of climate change, yet suffer from the  lack of effective remote sensing-based solutions \cite{burke2021using}. Foundation models for Earth monitoring have the potential to address many of these issues and substantially accelerate and enable the development of new remote sensing solutions for climate change.

\subsection{Increased accessibility}
Reducing the need for curating a large labeled dataset for each task could democratise access to the development of machine learning models for remote sensing, specifically for groups or organisations with limited budgets \cite{maskey2020advancing, Alemohammad2021}. In particular, foundation models may especially benefit non-profit organisations, academic universities, startups, and developing countries. It may also open opportunities for applications that were not previously profitable. Although we believe that increased accessibility to these models will have a largely net positive impact, we acknowledge that this accessibility may lead to unexpected applications with potentially negative impacts \cite{bommasani2021opportunities}. We also note that such models may have dual-use applications, where, for example, they may help oil and gas industries in their operations in ways that increase (or reduce) overall emissions.

\subsection{Emissions of large pre-trained models} 
Recent work has investigated emissions of large neural networks \cite{strubell2019energy, schwartz2020green, codecarbon, lacoste2019quantifying, patterson2021carbon}. Specifically, training a large transformer can emit 284 \tco when trained on computers using largely  fossil fuel energy (US national average) \cite{strubell2019energy}. When put in perspective with individual actions, such emissions are large---e.g., a roundtrip passenger flight from San Francisco to London is 2.8 \tco, about 100$\times$ smaller. However, the extensive reusability of pre-trained models and their potential for helping efforts to mitigate climate change~\cite{rolnick2019tackling} calls for a different perspective.

When evaluating new tools and systems, it is important to consider the likely net impact on emissions of both the creation and testing of the tool and its eventual deployment. For example, evaluating the performance of airborne methane sensing tools at emission levels commonly found in oil and gas operations can emit about 7 metric tonnes of methane, roughly 600 \tco equivalent using a 20-year global warming potential \cite{epa_greenhouse_2017}. However, in a single day of flying, such a single instrument can survey hundreds of sites, often identifying leaks for repair that emit well over 7 metric tonnes of methane per day \cite{johnson_airborne_2021}. Similarly, foundation models may significantly advance our ability to leverage enormous quantities of passively collected satellite data to massively reduce emissions, qualitatively advance our understanding of climate science, or improve our ability to adapt to climate change.

In sum, the potential benefits for climate change mitigation with improved Earth monitoring methods likely outweigh the emissions associated with foundation models. Moreover, various actions can be taken to reduce and mitigate emissions related to the training of your model \cite{lacoste2019quantifying}:
\begin{compactitem}
    \item Select data centers that are certified carbon neutral or largely powered by renewable energy, with good power usage effectiveness (PUE). Such measures can reduce emissions dramatically ~50$\times$ reduction in emissions \cite{lacoste2019quantifying}.
    \item Design your code development pipeline to minimize the number of computationally-intensive runs required, e.g. employ modular development and testing when possible.
    \item Make your code more efficient and sparsify your network when possible \cite{patterson2021carbon}. This can reduce emissions up to ~10-fold.
    \item Favour more energy-efficient hardware, e.g., TPUs or GPUs.
    \item Monitor~\cite{codecarbon} and report your emissions \cite{lacoste2019quantifying}. Better communication about climate change is fundamental for systemic changes. Better documentation will help other coders pick up where you left off, potentially bypassing some computationally intensive runs.
    \item Offset the cumulative emissions of your projects.
\end{compactitem}

\subsection{Fairness and biases}
Large language models are known to amplify and perpetuate biases \cite{bender2021dangers}. While this can lead to serious societal issues, we believe that biases in remote sensing models are likely to have much less impact. We do however anticipate potential biases and fairness issues. 

\mypar{Data coverage and resolution} Some satellites cover the whole Earth with standard spatial resolution and revisit rate (e.g., Sentinel-2 covers the whole Earth at 10-60 m/pixel resolution every 5 days). This makes imagery freely available uniformly across the planet. Other satellite data providers such as Maxar acquire images on-demand and have higher spatial resolution (up to 0.3m per pixel), but also have lower revisit rates and high costs. Some countries, such as New Zealand, freely provide aerial imagery with resolution up to 0.1m per pixel\footnote{\url{https://data.linz.govt.nz/}}. Finally, it is worth noting that cloudy seasons in some climates may limit data availability for some countries. Overall, while the coverage is fairly uniform, some regions have much higher coverage than others and money can be a limiting factor to access the data. This can lead to some level of biases and fairness issues.

\section{List of Downstream Tasks} \label{sec:tasks_table}

\begin{table*}[p][h]
    \begin{center}
    \begin{tabular}{|c|c|c|c|c|c|}
         \hline
         Name & Task & Sector & \# labels & Resolution & Spectral bands \\
         \hline
         \makecell{\href{http://registry.mlhub.earth/}{AgricultureVision} \\\cite{chiu_agriculture-vision_2020}} & \makecell{ Multi-class classification or \\segmentation of agricultural \\patterns important to \\farmers (e.g., planter \\skip or nutrient deficiency) \\in aerial images.} & Agriculture & 94,986 & 10cm & \makecell{RGB + \\near infrared}\\
         \hline
         \makecell{\href{https://github.com/GlobalHydrologyLab/AquaSat}{AquaSat}\\\cite{ross_aquasat_2019}} & \makecell{Per-pixel regression to \\predict water quality \\(e.g., total suspended \\sediments) in \\satellite images.}& Water quality& ~600,000& 30m& \makecell{Multispectral \\+ RGB}\\
         \hline
         \makecell{\href{http://registry.mlhub.earth/10.14279/depositonce-10149/}{CalMethane Survey}\\\cite{duren_californias_2019}} & \makecell{Methane plume \\classification} & Energy & ~60-1000 & 3m & Hyperspectral \\
         \hline
         \makecell{\href{https://github.com/FrontierDevelopmentLab/CUMULO}{CUMULO}\\\cite{zantedeschi2019cumulo}} & \makecell{Detecting clouds to \\reduce uncertainties in \\climate models} & Climate & 300,000 & 1km & \makecell{Hyperspectral \\(36-band)} \\
         \hline
         \makecell{\href{http://registry.mlhub.earth/10.34911/rdnt.d2ce8i/}{LandCoverNet}\\\cite{alemohammad_landcovernet_2020}} & \makecell{Segmentation via multispectral \\satellite imagery with \\annual land cover \\class per pixel} & Land use & \~2,000 & 10m & Multispectral \\
         \hline
         \makecell{\href{https://mediatum.ub.tum.de/1483140}{SEN12-FLOOD}\\\cite{rambour_flood_2020}} & \makecell{Image classification \\of multispectral and \\radar satellite imagery \\to identify flooded regions}& \makecell{Climate/\\Adaptation} & 5,567 & 10m & \makecell{Multispectral\\+SAR} \\
         \hline
         \makecell{\href{http://registry.mlhub.earth/10.34911/rdnt.xs53up/}{Tropical cyclone wind speed}\\\cite{Maskey2020}} & \makecell{Regression-based estimation \\of surface wind speed \\of tropical cyclones \\using satellite imagery} & Climate & 114,634 & 4km & \makecell{Single-band \\microwave} \\
         \hline
         \makecell{\href{https://github.com/kdmayer/3D-PV-Locator}{3D PV Locator}\\\cite{rausch2020enriched}} & \makecell{Classification and \\segmentation of solar \\panels in satellite \\imagery}& Energy & 100,000 & 10cm & RGB \\
         \hline
    \end{tabular}
    \caption{Example datasets for benchmarking climate-focused Erath monitoring foundation models. All listed datasets satisfy the criteria presented in Section~\ref{sec:tasks}}
    \label{tab:datasets}
    \end{center}
\end{table*}
\end{document}